\documentclass[letterpaper, 10 pt, conference]{ieeeconf}  
\IEEEoverridecommandlockouts                              

\usepackage{svg}
\usepackage{cite}
\usepackage{tabularx}
\usepackage{colortbl}
\usepackage{multirow}
\usepackage{multirow}
\usepackage{amsmath,amssymb,amsfonts}
\usepackage{mathtools, bigstrut}
\usepackage{graphicx}
\usepackage{textcomp}
\usepackage{wrapfig}

\usepackage[linesnumbered,ruled,lined]{algorithm2e}

\usepackage{xcolor}
\usepackage{soul}

\usepackage{multirow} 
\usepackage{makecell} 
\usepackage{booktabs} 
\usepackage{array} %
\usepackage{colortbl}

\usepackage{authblk}
\let\labelindent\relax
\usepackage{enumitem}
\usepackage[pagebackref]{hyperref}
\hypersetup{
    colorlinks=true,
    citecolor=ForestGreen,
    linkcolor=blue,
    filecolor=magenta,      
    urlcolor=cyan
}
\usepackage{breqn}
\usepackage{breqn}
\usepackage{algpseudocode}
\usepackage[font=footnotesize]{caption}
\usepackage{subcaption}
\usepackage{tcolorbox}
\usepackage{array}
\usepackage{makecell}
\usepackage{cuted}
\usepackage{epsfig}

\usepackage[english]{babel}
\usepackage{amsthm}

\usepackage{tabularray}
\definecolor{Silver}{rgb}{0.752,0.752,0.752}
\definecolor{Gallery}{rgb}{0.937,0.937,0.937}
\usepackage{adjustbox}

\allowdisplaybreaks

\title{\LARGE \bf
Crowd-FM: Learned Optimal Selection of Conditional Flow Matching-generated Trajectories for Crowd Navigation
}
\author{\vspace{-0.5mm}
Antareep Singha$^{2*}$, Laksh Nanwani$^{1*}$, Mathai Mathew P.$^{1}$, Samkit Jain$^{1}$, Phani Teja Singamaneni$^{4}$, \\Arun Kumar Singh$^{3}$, K. Madhava Krishna$^{1}$ 

\thanks{* Equal contribution.}
\thanks{$^{1}$ Robotics Research Center, IIIT Hyderabad, India. \textcolor{gray}{\scriptsize \{lakshanshul, mathewp8616\}@gmail.com, \{samkit.jain@students, mkrishna\}@iiit.ac.in}}
\thanks{$^{2}$ Nanyang Technological University, Singapore. \textcolor{gray}{\scriptsize antareep002@e.ntu.edu.sg}}
\thanks{$^{3}$ University of Tartu, Estonia. \textcolor{gray}{\scriptsize aks1812@gmail.com}}
\thanks{$^{4}$ Inria, Universit\'e de Lorraine, France. \textcolor{gray}{\scriptsize phaniteja.sp@gmail.com}}
\thanks{\textbf{Project Page}: \href{https://smart-wheelchair-rrc.github.io/crowdfm-webpage/}{https://smart-wheelchair-rrc.github.io/crowdfm-webpage/}}

\thanks{We acknowledge IHub-Data(Project:\textbf{M2-029}) for funding this work. It was also co-funded by the European Union and Estonian Research Council via Project:\textbf{TEM-TA101} and Grant:\textbf{PSG753} by Estonian Research Council.}}

\begin{document}

\maketitle
\thispagestyle{empty}
\pagestyle{empty}

\begin{strip}
\centering
\vspace{-9.0em}
\includegraphics[width=\linewidth]{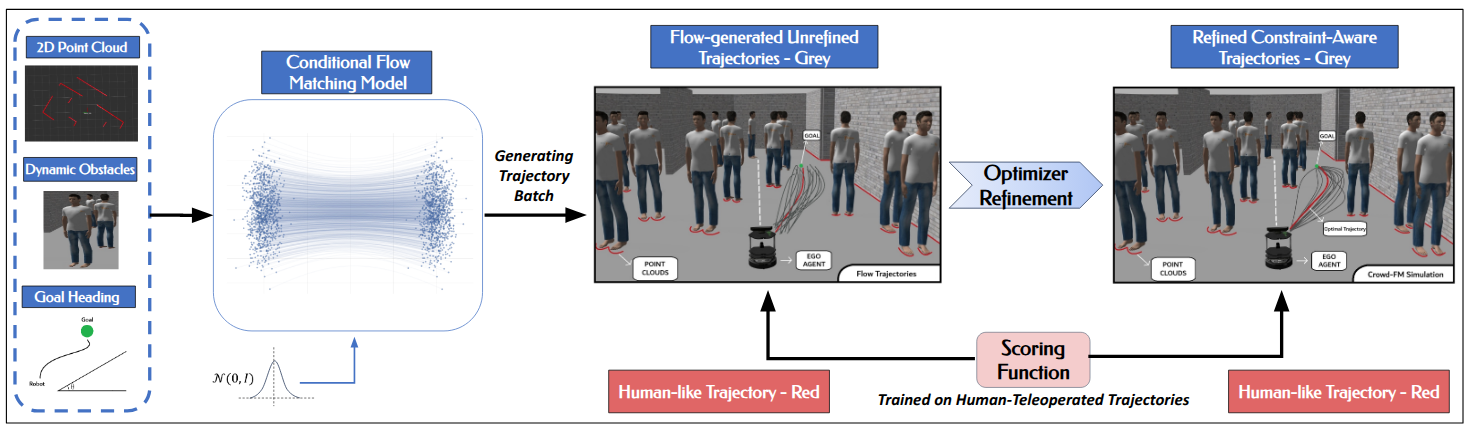}
\captionof{figure}{\footnotesize{Crowd-FM is a long-horizon local planner that is capable of rapidly generating collision-free trajectories in a batch. It takes in as input 2D point cloud data, dynamic obstacle positions and velocities, and the heading-to-goal angle. A Conditional Flow Matching model is trained to generate such trajectories conditioned on the input sensor data. Finally, the trajectories are refined using a Projection Optimizer\cite{rastgar2023priestprojectionguidedsamplingbased} to meet kinodynamic constraints. A separate Scoring Function is trained on Human Expert Trajectories, to enable selection of human-like trajectories from the ones generated by CFM.}}
\label{fig:fig_teaser}
\vspace{-0.4cm}
\end{strip}

\begin{abstract}
Safe and computationally efficient local planning for mobile robots in dense, unstructured human crowds remains a fundamental challenge. Moreover, ensuring that robot trajectories are similar to how a human moves will increase the acceptance of the robot in human environments. In this paper, we present Crowd-FM, a learning-based approach to address both safety and human-likeness challenges. Our approach has two novel components. First, we train a Conditional Flow-Matching (CFM) policy over a dataset of optimally controlled trajectories to learn a set of collision-free primitives that a robot can choose at any given scenario. The chosen optimal control solver can generate multi-modal collision-free trajectories, allowing the
CFM policy to learn a diverse set of maneuvers. Secondly, we learn a score function over a dataset of human demonstration trajectories that provides a human-likeness score for the flow primitives. At inference time, computing the optimal trajectory requires selecting the one with the highest score. Our approach improves the state-of-the-art by showing that our CFM policy alone can produce collision-free navigation with a higher success rate than existing learning-based baselines. Furthermore, when augmented with inference-time refinement, our approach can outperform even expensive optimisation-based planning approaches. Finally, we validate that our scoring network can select trajectories closer to the expert data than a manually designed cost function.\\
\end{abstract}

\vspace{-3mm}
\section{INTRODUCTION}
Mobile robots are still far from achieving effective trajectory modelling that enables them to operate in scenarios characterized by extreme dynamic complexity. Such scenarios are difficult to navigate even for humans, and in these situations, simply prioritizing safety above all else becomes the norm. Classical planners, with their mathematically rigorous models, are effective in ensuring safety but often fail in environments with high uncertainty and variability, as seen in velocity-obstacle–based approaches such as RVO \cite{4543489}. In contrast, recent approaches based on generative modelling, such as VQ-VAEs \cite{11128132,idoko2024learningsamplingdistributionsafety} or Reinforcement Learning \cite{Xie_2023,chen2019crowdnav,DBLP:journals/corr/abs-2002-03038}, attempt to capture complex behaviors but suffer from limitations in scalability, sample efficiency, and generalization. Furthermore, much of the existing literature demonstrates success only in relatively simple, low-interaction settings \cite{chen2020relational}, while performance degrades substantially in densely interactive environments common in real-world applications.

The core concept of Crowd-FM is to reframe the planning problem. Instead of searching for a single optimal path, it first generates a rich distribution of plausible future trajectories using Conditional Flow Matching \cite{lipman2023flow,tong2024improvinggeneralizingflowbasedgenerative}, which provides smoother and more expressive trajectory representations compared to discrete latent methods such as VQ-VAEs. A learned Scoring Function then selects the optimal one from the pool of Flow-generated trajectories, achieving a balance between robustness and efficiency.

While some works condition Flow Matching on 3D point clouds \cite{chisari2024learning} or depth priors from RGB-D images \cite{gode2025flownav}, our approach uses 2D LiDAR data complemented by dynamic obstacle states. These states can be obtained from lightweight trackers \cite{legtracker} \cite{LV-DOT} in the real world, avoiding the computational overhead of high-dimensional visual inputs. This design creates an efficient balance between representation richness and real-time feasibility, making Crowd-FM readily deployable on resource-constrained mobile platforms.


\section{RELATED WORKS}

\textbf{Classical and Optimization-Based Navigation: }
Early approaches to robot navigation in crowded environments were grounded in geometric formulations such as Velocity Obstacles and Reciprocal Velocity Obstacles (RVO) \cite{4543489}, which provide mathematically rigorous methods for collision avoidance. Time-Elastic-Band-based\cite{singamaneni2020hateb} planners like CoHAN\cite{singamaneni2021human, singamaneni2022watch} have also proven to be human-aware in moderately crowded settings. Sampling-based optimization has been explored, with methods such as PRIEST \cite{rastgar2023priestprojectionguidedsamplingbased} augmenting trajectory sampling with projection-based feasibility checking. While effective at ensuring robot safety, these approaches often become overly conservative in highly dynamic, interactive environments.
\newline
\textbf{Reinforcement Learning for Navigation: }
Deep reinforcement learning (DRL) has been extensively studied for crowd navigation. Works such as CrowdNav \cite{chen2019crowdnav} and its graph-based extensions \cite{chen2020relational} introduced policies that account for human motion and interactions. DRL-VO \cite{Xie_2023} integrated velocity obstacle constraints into the DRL framework, while DenseCAvoid \cite{DBLP:journals/corr/abs-2002-03038} demonstrated anticipatory behaviors in dense crowds. Although these methods show promise, their performance degrades in highly dynamic scenes, and training often requires large-scale data with careful reward shaping.
\newline
\textbf{Generative Models for Trajectory Planning: }
Generative modelling has recently emerged as a promising alternative to reinforcement and classical methods. Diffusion-based approaches such as NoMaD \cite{10610665} learn goal-conditioned distributions for long-horizon planning, while VQ-VAE–based methods like CrowdSurfer \cite{11128132} and variations with differentiable safety filters \cite{idoko2024learningsamplingdistributionsafety} have achieved strong results in dense navigation tasks. However, these models suffer from limitations such as computational overhead (diffusion) or due to latent bottlenecks which quantize away detail (VQ-VAE).
\newline
\textbf{Flow-Based Generative Models: }Flow-based methods provide an alternative for distribution learning. Lipman et al. \cite{lipman2023flow} introduced Flow Matching as a stable and efficient training paradigm, with subsequent work improving generalization via minibatch optimal transport \cite{tong2024improvinggeneralizingflowbasedgenerative}. Applications of Conditional Flow Matching have been demonstrated in robotic manipulation from pointclouds \cite{chisari2024learning} and navigation with depth priors \cite{gode2025flownav}. In this work, \textbf{we extend these ideas to dense crowd navigation}, where Flow Matching enables robust trajectory distribution learning, combined with a learned scoring function to boost success.

\section{METHODOLOGY}
\vspace{-0.1cm}

In this section, we set up the crowd navigation problem and its input space representation, subsequently describing how we solve it. We describe Conditional Flow Matching(CFM), the core generative part of Crowd-FM, and how our CFM model architecture is set up. We then explain how we design a learned Scoring Function to choose the optimal trajectory from a batch of Flow-generated candidates.

\subsection{Trajectory Parameterization}
\label{Traj_param}

The primary objective is to generate a continuous and dynamically feasible trajectory defined by its coordinates $x(t),y(t)$ over a time horizon. The trajectory generation process is conditioned on a context vector, $C$, which encapsulates high-dimensional sensory inputs (environmental context and goal).
Directly learning a mapping from the context $C$ to the infinite-dimensional space of continuous functions is intractable. To achieve this, we parameterize the continuous robot trajectories using a Bernstein polynomial of order $n(10)$, uniquely defined by a set of $n+1$ control points, over the time horizon $t \in [t_1, t_{n}]$. This approach effectively transcribes the infinite-dimensional optimal control problem into a finite-dimensional nonlinear programming problem, which is more amenable to machine learning algorithms.

The parameterization is done in the following manner:

\begin{align}
    \begin{bmatrix}
    x(t_{1}) \\ \vdots \\ x(t_n)
    \end{bmatrix}^{T} \hspace{-0.3cm} = \mathbf{P} \hspace{0.1cm}\bold{c}_{x} ,\hspace{-0.1cm}
     \begin{bmatrix}
    y(t_{1})\\ \vdots\\ y(t_n)
    \end{bmatrix}^{T} \hspace{-0.2cm}= \mathbf{P} \bold{c}_{y}, \mathbf{W} = \begin{bmatrix}
        \mathbf{P} & \mathbf{0}\\
        \mathbf{0} & \mathbf{P}
    \end{bmatrix}\label{parametrized}
\end{align}

\begin{figure*}[h!]
    \centering
    \includegraphics[width=0.90\textwidth]{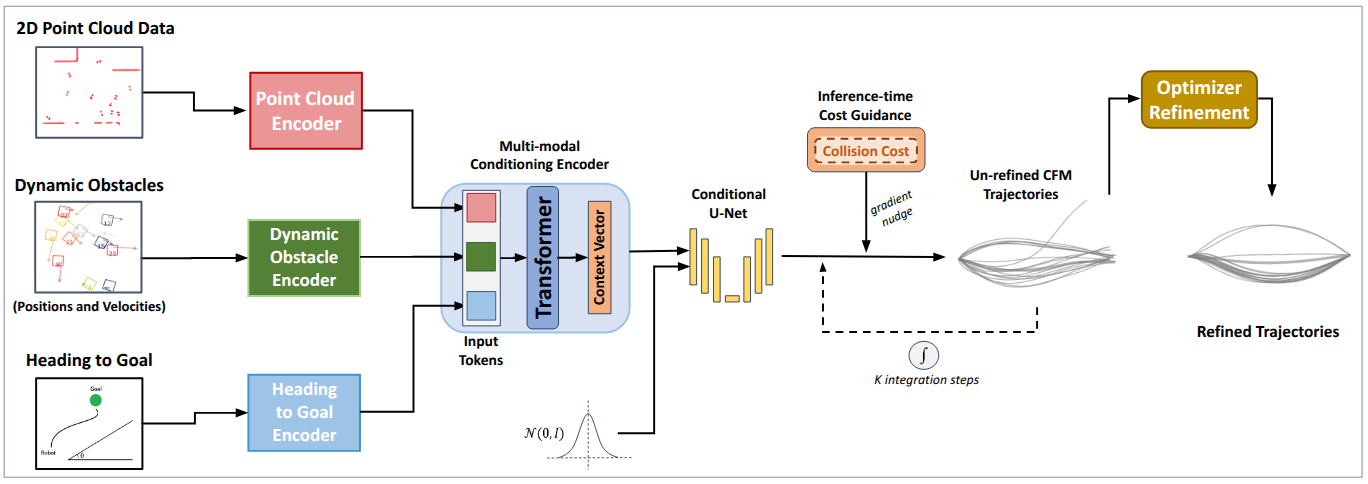}
    \caption{\footnotesize{The Conditional Flow Matching model is used to learn a multi-modal distribution of collision-free trajectories in terms of Bernstein coefficients to ensure that the reconstructed trajectories have higher continuity and differentiability. The model is conditioned on a Transformer-based input space encoder that takes into account the environmental context at every timestep. The inference-time integration is implicitly guided by a collision cost term encouraging collision-free generations. Finally, the generated trajectories are refined using a single optimization step\cite{rastgar2023priestprojectionguidedsamplingbased} to satisfy kinodynamic constraints. 
    }
    }
    \label{flow}
    \vspace{-0.5cm}
\end{figure*}

\noindent where the matrix $\mathbf{P}$ is a matrix formed by time-dependent polynomial basis functions. The vectors $\mathbf{c}_x, \mathbf{c}_y$ are the coefficients attached to the individual basis functions. We can represent the derivatives like $\dot{x}(t), \ddot{x}(t), \dot{y}(t), \ddot{y}(t) $ in a similar manner as \eqref{parametrized} using $\dot{\mathbf{P}}$ and $\ddot{\mathbf{P}}$. The combined trajectory representation is then given as $\boldsymbol{\xi} = \begin{bmatrix} \mathbf{c}_x & \mathbf{c}_y \end{bmatrix}^{T}$.

Our trajectory representation follows previous motion planning works like \cite{11128132}, \cite{idoko2024learningsamplingdistributionsafety}, and \cite{rastgar2023priestprojectionguidedsamplingbased}.
By choosing to generate Bernstein control points rather than a sequence of waypoints, the generative model is not burdened with learning the concepts of smoothness or continuity from data since the mathematical structure of the Bernstein representation guarantees these priors.  However, ensuring the resulting trajectories adhere to the robot's dynamic constraints requires further refinement, which is addressed in \ref{refinement}. Although our choice of Bernstein polynomials is motivated by the same advantages demonstrated in recent diffusion-based works such as GPD \cite{srikanth2025gpdguidedpolynomialdiffusion}, the way these representations are used differs substantially. GPD leverages diffusion over Bernstein coefficients with a guided denoising process and a trajectory stitching procedure to ensure feasibility. In contrast, Crowd-FM uses CFM to directly learn the transformation to Bernstein coefficients via ODE integration, enabling faster and more stable inference.



\subsection{Flow Matching Objective}

CFM learns a time-dependent vector field, $u_\tau(\boldsymbol{\xi}),$ that transports samples from a simple, known prior distribution, \( q_0(\boldsymbol{\xi}) \) (a standard normal distribution \(\mathcal{N}(0, I)\)), to a complex target data distribution, \( q_1(\boldsymbol{\xi}) \). This transport is achieved by integrating the ODE
\[\\d\boldsymbol{\xi} = u_\tau(\boldsymbol{\xi})d\tau\] from an \textbf{``artificial time''} \( \tau = 0 \) to \( \tau = 1 \). It is worth noting that this ``artificial time'' is distinct from the time $t$ used in \ref{Traj_param}. A neural network, \( v_\theta(\boldsymbol{\xi}, \tau, C) \), with $C$ as the condition(context vector), is trained to approximate this true vector field. A key advantage of CFM is its use of a regression-based objective that is tractable and avoids the need for costly ODE simulation during the training phase.

As discussed in subsection \ref{Traj_param}, we train the CFM model directly in the space of Bernstein control points, represented by the vector \(\boldsymbol \xi \in \mathbb{R}^{2(n+1)}.\) The prior distribution, \( q_0(\boldsymbol{\xi}) \), is a standard multivariate Gaussian, while the target distribution, \( q_1(\boldsymbol{\xi}) \), is the empirical distribution of control points extracted by fitting Bernstein polynomials to ground truth trajectories in the training dataset.
\newline
Following the independent coupling strategy employed in recent works like \cite{gode2025flownav}, a conditional probability path \(p_\tau(\boldsymbol{\xi}|\boldsymbol{\xi_0}, \boldsymbol{\xi_1})\)
and a corresponding vector field \(u_\tau(\boldsymbol{\xi}|\boldsymbol{\xi_0}, \boldsymbol{\xi_1})\) is defined for any pair of samples \(\boldsymbol{\xi_0} \sim q_0(\boldsymbol{\xi})\) and \(\boldsymbol{\xi_1} \sim q_1(\boldsymbol{\xi}).\) This path is a simple linear interpolation in the high-dimensional space of control points: 
\[\boldsymbol{\xi}_\tau = (1 - \tau)\boldsymbol{\xi_0} + \tau\boldsymbol{\xi_1}\]

The vector field that generates this path is constant with respect to both time and position along the path, pointing directly from the noise sample to the data sample:

\[
u_\tau(\boldsymbol{\xi}|\boldsymbol{\xi_0}, \boldsymbol{\xi_1}) = \boldsymbol{\xi_1} - \boldsymbol{\xi_0}
\]
Consequently, the neural network $v_\theta$ is trained on a manifold of physically valid and meaningful trajectory representations at every intermediate flow time $\tau$. 

For our problem, the neural network, \( v_\theta(\boldsymbol{\xi_\tau}, \tau, C) \), is designed to predict the target vector field \( u_\tau = \boldsymbol{\xi_1} - \boldsymbol{\xi_0} \). The model is trained by minimizing the Conditional Flow Matching loss, which takes the form of a simple mean squared error objective:
\[
\mathcal{L}_{\mathrm{CFM}}(\theta) = \mathbb{E}_{\tau \sim U(0, 1), \boldsymbol{\xi_0} \sim q_0, \boldsymbol{\xi_1} \sim q_1, C}\]
\[\left[ \| v_\theta((1 - \tau)\boldsymbol{\xi_0} + \tau\boldsymbol{\xi_1}, \tau, C) - (\boldsymbol{\xi_1} - \boldsymbol{\xi_0}) \|^2 \right]
\]

This objective directly regresses the output of the learned vector field onto the constant-velocity field that connects a random noise sample to a target data sample, conditioned on the environmental context.

\vspace{-0.1cm}

\subsection{Flow Matching Model Architecture}

This subsection details the Conditional Flow Matching architecture (Fig. \ref{flow}) adapted to learn the complex multi-modal distribution of Bernstein control points that define expert trajectories.

The architecture comprises two primary components: a multi-modal input space encoder that processes the environmental and goal information into a context vector $C$, and a conditional U-Net that approximates the vector field $v_\theta$. Subsequently, Inference-time Cost Guidance and Optimizer refinement are added to the pipeline, as discussed in subsections \ref{guidance} and \ref{refinement} respectively, to ensure safety and smoothness at all times.

\subsubsection{Multi-modal Conditioning Encoder}

The multi-modal conditioning encoder processes static obstacles, dynamic obstacles, and heading to goal angles at every timestep through specialized encoders before fusing them using a self-attention mechanism.

\paragraph{Point Cloud Encoder} For a point cloud input \( X_{\mathrm{pcd}} \in \mathbb{R}^{N_{\mathrm{pts}} \times 2} \), where \( N_{\mathrm{pts}} \) is the maximum number of point-cloud points, a feature extractor \( f_{\mathrm{pcd}} \) composed of a series of 1D convolutions and ReLU activations maps each point to a high-dimensional feature vector. A global max-pooling operation is then applied across all points to yield a permutation-invariant global feature vector, \( c_{\mathrm{pcd}} = \max_{i=1...N_{\mathrm{pts}}} f_{\mathrm{pcd}}(X_{\mathrm{pcd}}) \). This vector is processed by a multi-layer perceptron (MLP), \( g_{\mathrm{pcd}} \), to produce the final static embedding, \( C_{\mathrm{static}} = g_{\mathrm{pcd}}(c_{\mathrm{pcd}}) \).
\paragraph{Dynamic Obstacle Encoder} The Dynamic Obstacle information is compiled into a specialized tensor for the model to work with. For an input dynamic obstacle tensor \( X_{\mathrm{dyn}} \in \mathbb{R}^{N_{\mathrm{obs}} \times 4} \), representing $N_{\mathrm{obs}}$, each with a 4D kinematic state(position and velocity), a dynamic obstacle encoder \( f_{\mathrm{dyn}} \) processes them into a high-dimensional feature vector. An MLP converts obstacle features into high-dimensional embeddings, which are augmented with positional embeddings. Subsequently, a minimal Transformer Encoder with 4 heads and w layers employs self-attention to model interactions among the obstacles. Finally, max-pooling aggregates these features into a single vector, which a final MLP processes to produce the dynamic context embedding$(D=512)$, $C_{\mathrm{dynamic}}$.
\paragraph{Goal Encoder} The navigation goal is provided as a 2D heading vector, \( X_{\mathrm{goal}} \in \mathbb{R}^2 \). This vector is processed by a dedicated MLP, \( g_{\mathrm{goal}} \), to produce a goal embedding: \( C_{\mathrm{goal}} = g_{\mathrm{goal}}(\mathrm{X}_{\mathrm{goal}}) \).
\paragraph{Multi-modal Fusion}The individual embeddings, \( C_{\mathrm{static}} \), \( C_{\mathrm{dyn}} \), and \( C_{\mathrm{goal}} \), are treated as tokens for a fusion module. The tokens are concatenated to form an input sequence, which is then processed by \( F_{\mathrm{fusion}} \), a Transformer Encoder with 8 heads and 3 layers. The Transformer Encoder uses self-attention to produce the final context, a sequence of fused tokens \( C = F_{\mathrm{fusion}}(S) \). This sequence captures the complex interplay between the robot's environment, other moving agents, and its objective. This context is directly used to condition the 1D U-Net as explained below.

\vspace{0.2cm}
\subsubsection{Conditional U-Net}
The core of Crowd-FM is the flow prediction network that is implemented as a 1D U-Net. This network is designed to predict the vector field that transports the control points from the prior to the data distribution.

The U-Net takes three inputs:

\begin{enumerate}
    \item The interpolated control point vector \( \boldsymbol{\xi_\tau} \in \mathbb{R}^{2(n+1)} \).
    \item The scalar flow time \( \tau \in [0,1] \).
    \item The sequence of context tokens \( C \) from the conditioning encoder.
\end{enumerate}

The U-Net architecture consists of a symmetric encoder-decoder structure with skip connections. The encoder progressively downsamples the 1D representation of the trajectory control points, and the decoder upsamples this representation, integrating information from the corresponding encoder layers via skip connections. The flow time \( \tau \) is transformed into a sinusoidal time embedding and added to the intermediate features at each resolution level. The context \( C \) is injected into the network within the U-Net's residual blocks, allowing the model to condition the predicted flow on the specific environmental and goal context. The final output of the network is the predicted vector field \( v_\theta(\boldsymbol{\xi_\tau}, \tau, C) \), which has the same dimension as the input \( \boldsymbol{\xi_\tau} \).
\vspace{-0.1cm}
\subsection{Inference-time Corrections} 

\noindent We perform two kinds of inference-time modifications to the flow predicted trajectories. First, we embed a guidance step into the flow integration process. Second, we use a more powerful projection optimizer on the flow output to further push it towards feasible regions.

\subsubsection{Guided Integration Process}
\label{guidance}
To encourage the integration process to generate collision-free trajectories at run-time in dense crowds, we apply an inference-time collision cost guidance, quite similar to how \cite{Liu_2023_CVPR} implements gradient-based control for image generation. The cost is formulated as: 
\[
\mathcal{L}_{\text{collision}} = \frac{1}{N} \sum_{k=1}^{N_{\text{waypoints}}} \max\left(0, d_{\text{safe}}^2 - \min_{m=1}^{N_{\text{points}}} \|\mathbf{p}_k - \mathbf{o}_m\|_2^2\right)
\]
where:
\begin{itemize}
    \item \( \mathbf{p}_k(x(t),y(t)) \) is the \( k \)-th waypoint along the trajectory(transformed from coefficients)
    \item \( \mathbf{o}_m \) is the \( m \)-th obstacle point
    \item \( d_{\text{safe}}\) is the safety margin
    \item \( N_{\text{points}} \) is the number of obstacle points
\end{itemize}

It is important to note that $\mathbf{p}_k$ is a function of $\boldsymbol{\xi_\tau}$, and is implied that:
$$
\mathcal{L}_{\text{collision}} = f(\mathbf{p}_k), \quad \text{and}
$$
$$
\quad \mathbf{p}_k = g(\boldsymbol{\xi_\tau}) \implies \mathcal{L}_{\text{collision}} = f(g(\boldsymbol{\xi_\tau}))
$$
The gradient of this function, \( \nabla \mathcal{L}_{\text{collision}} \), is a vector that points in the direction of the steepest ascent of the cost. Consequently, the negative gradient, \( -\nabla \mathcal{L}_{\text{collision}} \), points towards more desirable, lower-cost states.

We recall from the Flow Matching Objective that the ODE is formulated as:
\[
\frac{d\boldsymbol{\xi_\tau}}{d\tau} = v(\boldsymbol{\xi_\tau}, \tau, C)
\]

To encourage collision-free generation, we modify the original ODE by adding this negative gradient term to the learned vector field. The new, guided ODE is defined as:

\[
\frac{d\boldsymbol{\xi_\tau}}{d\tau} = v(\boldsymbol{\xi_\tau}, \tau, C) - \lambda \cdot \nabla_{\boldsymbol{\xi_\tau}} \mathcal{L}_{collision}(\boldsymbol{\xi}_\tau)
\]

where \( \lambda \) is a scalar guidance scale hyperparameter that controls the strength of the corrective guidance. By solving this modified ODE, the state \( \boldsymbol{\xi_\tau} \) evolves according to a composite vector field. It simultaneously follows the learned data manifold (via \( v(\boldsymbol{\xi_\tau}, \tau) \)) and descends the gradient of the cost function (via \(-\lambda \cdot \nabla_{\boldsymbol{\xi_\tau}} \mathcal{L}_{\text{collision}}(\boldsymbol{\xi}_\tau)\)).
\vspace{0.2cm}

\subsubsection{Optimizer Refinement (Projection Optimizer)}
\label{refinement}
The final flow trajectories are further refined by the projection optimizer of \cite{rastgar2023priestprojectionguidedsamplingbased}, which produces the closest trajectories to the flow prediction that is also collision-free.


\subsection{Learned Scoring Function}
The flow model is designed to output a roll-out of several trajectories, but one of them must be selected for traversal. This can be done in multiple ways, including a pre-defined cost-based selection where the trajectory with the lowest overall cost is selected. However, more often than not, this yields trajectories that deviate from human behavior. To account for human-like behavior in Crowd-FM, we move towards a Trajectory Selector that inherently learns how to choose trajectories based on expert human demonstrations. In this subsection, we explain how we train a learnable scoring function based on the dynamic environmental context to imitate expert demonstrations. The scoring function learns a mapping \(S(\boldsymbol{\xi}, C) \) of a candidate trajectory $\boldsymbol{\xi}$ and the current context vector $C$ to a scalar score, enabling the selection of the best trajectory.

\begin{figure}[ht]
    \centering
    \includegraphics[width=0.9\linewidth]{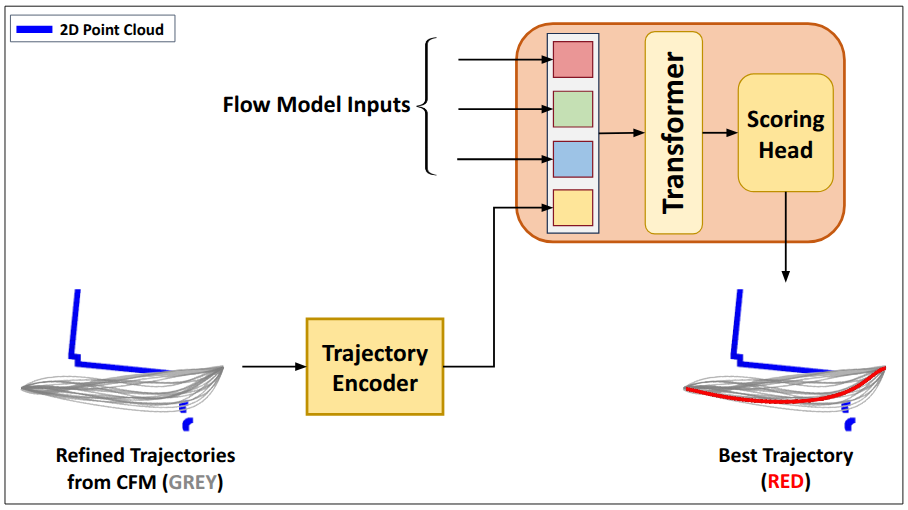}
    \caption{\footnotesize{The learned scoring function has a similar input space representation as the Flow model. It borrows the input encoders from the Flow model architecture, namely the Point Cloud, Dynamic Obstacles, and the Goal Encoders. Additionally, the Flow-generated trajectories are encoded and tokenized as the input to a Transformer Encoder. The transformer output is passed to the Scoring Head to output raw scores for each trajectory generated.}}
    \label{fig_score}
    \vspace{-0.4cm}
\end{figure}

\subsubsection{Scoring Function Architecture}
The scorer is a Transformer-based architecture that jointly reasons over the environmental context and a set of $K$ candidate trajectories generated at run-time.
\begin{itemize}
    \item \textit{Input Representation:} The scorer takes as input the same static obstacle, dynamic obstacle, and goal heading information as the flow model. Additionally, it takes a batch of $K$ flow-generated candidate trajectories, \(\{\boldsymbol{\xi}_1, ..., \boldsymbol{\xi}_K\}\)(each trajectory is represented by Bernstein control points).
    \item \textit{Input Space Encoders:} The scorer uses the same encoder modules as the flow model to produce context embeddings \( C_{\mathrm{static}}, C_{\mathrm{dyn}} \), and \( C_{\mathrm{goal}} \). A separate Trajectory Encoder is introduced to process the candidate trajectories. For each candidate \( \boldsymbol{\xi}_k \), its control points are passed through a 1D convolutional network followed by mean pooling and an MLP to produce a trajectory embedding \( C_{\mathrm{traj},k} \).
    \item \textit{Multi-modal Fusion and Scoring:} To differentiate between the sources of information, learnable modality-type embeddings (\( E_{\mathrm{static}}, E_{\mathrm{dyn}}, E_{\mathrm{goal}}, E_{\mathrm{traj}} \)) are added to their respective context embeddings. The context embeddings are then concatenated with the set of \( K \) trajectory embeddings to form a single sequence of tokens:
    \vspace{-0.2cm}
    
    \[
    S_{\mathrm{in}} = [C_{\mathrm{traj},1} + E_{\mathrm{traj}}, ..., C_{\mathrm{traj},K} + E_{\mathrm{traj}}, C_{\mathrm{static}}\]
    \[  + E_{\mathrm{static}}, C_{\mathrm{dyn}} + E_{\mathrm{dyn}}, C_{\mathrm{goal}} + E_{\mathrm{goal}}]
    \]
    \vspace{-0.2cm}
    
    This sequence is processed by a Transformer Encoder with 8 heads and 4 layers, allowing each trajectory token to attend to all other trajectory tokens and the full environmental context. The output tokens corresponding to the trajectories, \(\{\hat{C}_{\mathrm{traj},1}, \ldots, \hat{C}_{\mathrm{traj}, K}\}\), are then passed through a final MLP scoring head, which outputs a single scalar logit for each trajectory, producing the raw scores \(\{\mathrm{score}_1, \ldots, \mathrm{score}_K\}\).
\end{itemize}
\vspace{0.5mm}

\subsubsection{Training Objective}
The scorer is trained to identify which of the generated candidate trajectories is \textbf{``best''} with respect to an expert demonstration. Training the Scoring Function follows a similar trajectory as the final inference. A frozen, pre-trained flow model generates a set of \(K\) candidate trajectory coefficients that are optimizer refined and then converted to trajectories \(\{P_k\}_{k=1}^{K}\).\\
Given an expert trajectory \(P_{\mathrm{expert}}\) from our custom dataset of the ego-agent navigating in dense crowds, the ground truth label is determined by finding the candidate trajectory that is closest to the expert in Euclidean space. Let \(j\) be the index of this closest candidate:
\vspace{-0.1cm}
\[
j = \arg \min_{k \in \{1, \ldots, K\}} \| P_k(t) - P_{\mathrm{expert}}(t) \|_2
\]

The Scoring Function is trained using a cross-entropy loss, treating the problem as a K-class classification task, where the target labels are dynamically generated for each training instance. This objective trains the scorer to assign the highest logit to the candidate that best imitates the expert's path. During inference, it is sufficient to select the trajectory index with the highest score.\\
The loss is regularized by the cost from the optimizer discussed in section \ref{refinement}, encouraging the model to favor trajectories that are not only close to the expert but also have low optimization costs. The final loss is:
\vspace{-0.1cm}
\[
\mathcal{L}_{\mathrm{scorer}} = \text{CrossEntropy}(\mathrm{scores}, j) + \lambda \cdot \mathrm{mean}(\mathrm{cost}_{\mathrm{optimizer}})
\]

where \(\lambda\) is a hyperparameter balancing the two loss components.


For every set of \(K\) candidate trajectories, we define the ground truth on the fly by identifying the candidate that has the minimum Euclidean distance to the expert demonstration. The index of this ``closest'' candidate becomes the target label for the cross-entropy function. 

Therefore, while the absolute identity of any given index changes between training steps, the relative correspondence between the scores and the dynamically generated target is preserved, ensuring that the cross-entropy loss correctly penalizes the model for failing to assign the highest score to the geometrically optimal trajectory, as seen in Fig. \ref{fig_score}.

\subsection{Data Collection}


A critical factor in the performance of Crowd-FM is the quality of expert demonstrations used for training. However, existing public datasets for crowd navigation are typically limited to sparse interactions, simple layouts, or reactive behaviors, and therefore do not capture the dense, high-interaction settings needed for effective deployment. Moreover, they rarely emphasize smooth, non-freezing trajectories or provide data in a form readily compatible with Bernstein polynomial representations. To overcome these limitations, we curated a dataset specifically tailored to the requirements of Crowd-FM, combining both simulation and real-world scenarios.

\begin{table*}[t]
\centering
\renewcommand{\arraystretch}{1.2}
\begin{tabular}{|c|c|cc|cc|cc|cc|}
\hline
\textbf{World} & \textbf{Number of runs} & \multicolumn{2}{c|}{\textbf{DRL-VO}} & \multicolumn{2}{c|}{\textbf{CoHAN 2.0}} & \multicolumn{2}{c|}{\textbf{Crowdsurfer}} & \multicolumn{2}{c|}{\textbf{Crowd-FM(Ours)}} \\
\cline{3-10}
& & \textbf{Succ. runs} & \textbf{Rate} & \textbf{Succ. runs} & \textbf{Rate} & \textbf{Succ. runs} & \textbf{Rate} & \textbf{Succ. runs} & \textbf{Rate} \\
\hline
Cumberland & 15 & 6 & 0.40 & 10 & 0.67 & 11 & 0.73 & \cellcolor{green!25}\textbf{13} & \cellcolor{green!25}\textbf{0.87} \\
\hline
Lobby World & 18 & 11 & 0.61 & 14 & 0.78 & \cellcolor{green!25}\textbf{17} & \cellcolor{green!25}\textbf{0.94} & 16 & 0.89 \\
\hline
Freiburg & 15 & 8 & 0.53 & 9 & 0.60 & 11 & 0.73 & \cellcolor{green!25}\textbf{12} & \cellcolor{green!25}\textbf{0.80} \\
\hline
\end{tabular}
\vspace{-0.1cm}
\caption{\footnotesize{Comparison of navigation success rates across different planners and environments with \textbf{35 dynamic agents}. \textit{Note: CoHAN 2.0 success rates could be a little higher with more extensive parameter tuning.}}}
\label{tab:success_35}
\vspace{-0.8cm}
\vspace{1em}

\end{table*}

Data collection was performed using the Barn environment to capture challenging static obstacle layouts and PEDSIM to model dense dynamic crowds. The dataset consists of around five hours of navigation data using both the Jackal and Turtlebot2 robots. Two sources of trajectories were included:

\begin{enumerate}
    \item \textit{Manually teleoperated trajectories}: Collected in both simulation and real-world deployments, these are used to train the Scoring Function.
    \item \textit{Post-processed trajectories}: Refined with \cite{rastgar2023priestprojectionguidedsamplingbased} to ensure dynamics and consistency with the Bernstein polynomial representations employed for training the CFM.

\end{enumerate}

For training, the expert trajectories are parameterized with Bernstein polynomials, converting continuous paths into compact sets of control points. During inference, Crowd-FM is conditioned on three inputs: the 2D LiDAR point cloud, the heading angle to the global goal, and a tensor of dynamic obstacle states $(x, y, v_x, v_y)$.

\section{VALIDATION AND BENCHMARKING}


In this section, we detail our extensive evaluation of Crowd-FM. We show qualitative results, conduct quantitative benchmarking, and perform ablation studies in challenging simulated environments, comparing our method against three leading baselines: CrowdSurfer\cite{11128132}, DRL-VO\cite{Xie_2023}, and CoHAN2.0\cite{singamaneni2021human}. We then validate the practical feasibility of our pipeline in real-world settings.

\subsection{Qualitative Analysis}

This sub-section deals with the performance of the entire Crowd-FM pipeline across different PEDSIM environments:

\begin{itemize}
    \item \textit{Smooth and dynamically feasible trajectories:} Bernstein polynomial parameterization, coupled with Flow Matching, enforces continuity. This results in smooth trajectories, avoiding the jittery behavior observed in baseline planners.
    
    \item \textit{Robust reactivity to dynamic obstacles:} Flow-generated trajectories adapt quickly to the motion of surrounding agents, allowing the robot to maneuver without freezing, a common failure mode for VQ-VAE, optimization, and DRL-based methods.
    
    \item \textit{Diversity of options:} Unlike VQ-VAE, which tends to collapse onto a limited set of discrete maneuver templates, Crowd-FM generates a rich spectrum of candidate paths. As illustrated in Fig. \ref {fig_qual}, Crowd-FM proposes multiple feasible routes around obstacles, while CrowdSurfer's trajectories show a collapse to a unimodal distribution, highlighting the lack of diversity from its VQ-VAE priors when compared to the expressive representations learned by Flow Matching.
    
    \item \textit{Effective trajectory selection:} The learned Scoring Function leverages this diversity by consistently choosing the safest and most efficient option among Flow candidates. This selection mechanism allows Crowd-FM to combine exploration (via Flow) with refinement (via scoring), improving both safety and efficiency.
    
\begin{figure}[ht]
    \vspace{-0.2cm}
    \centering
    \includegraphics[width=0.7\linewidth]{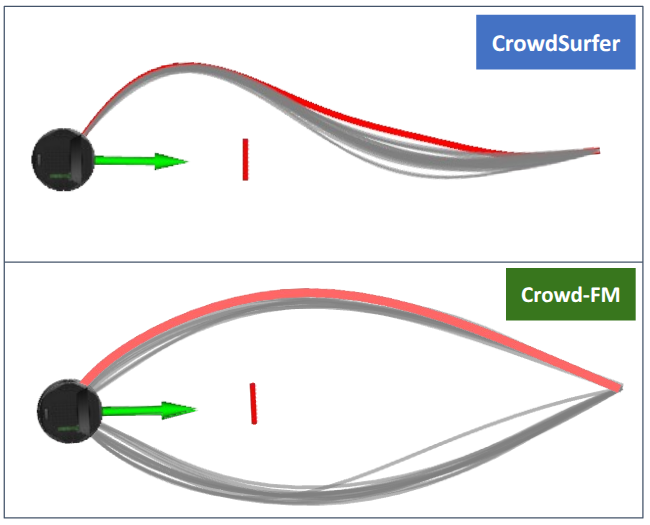}
    \caption{\footnotesize{\textit{Diversity of Trajectories: } The small red point cloud in front of the robot is the obstacle to be avoided. The upper image shows the trajectories generated by CrowdSurfer. The lower image shows the trajectories generated by Crowd-FM. The trajectories generated using Crowd-FM have more diverse outputs and can generate safe paths in multiple directions. The red trajectory is the best trajectory selected by both planners.}}
    \label{fig_qual}
    \vspace{-0.55cm}
\end{figure}

\end{itemize}
\subsection{Quantitative Analysis}

Crowd-FM sets a new standard for performance in crowd navigation, outperforming all baselines in nearly every metric across standard PEDSIM\cite{gloor_pedsim_2016} environments (Table \ref{tab:success_35}). While the performance of optimization-based planners like CoHAN 2.0 could potentially be improved through more exhaustive, environment-specific parameter tuning, Crowd-FM remains fundamentally superior in real-time inference efficiency, overall time-to-goal efficiency, and its inherent ability to prevent freezing behaviors in dense scenarios. This particularly improves upon CrowdSurfer, despite their shared use of generative priors. We stress-test both these pipelines in more challenging environments and share our results in Table \ref{crowdfm-v-crowdsurfer}.

This improvement stems from two pillars: (1) our CFM model's near-perfect learning of collision avoidance from data, and (2) exceptional computational speed (refer Sec.~\ref{sec:runtime}). The following subsections are dedicated to ablation studies, especially focusing on the effectiveness of the vanilla CFM-generated trajectories, followed by the runtime analysis of our pipeline.

\subsubsection{\textbf{Ablation Study without Optimizer Refinement}}

Here, we test Crowd-FM without the Optimizer refinement step by comparing the trajectories generated by the Vanilla-CFM with another baseline (DRL-VO) and see if the Conditional Flow Matching models the input data distribution well.
As is evident from Table \ref{tab:vanilla_results}, vanilla CFM alone is able to beat a leading RL-based baseline, DRL-VO, by a fair margin. This shows that CFM alone can model collision avoidance, especially in dense crowds. Qualitative examples of vanilla-CFM trajectories can be found in Fig. \ref{fig_px}.
\begin{table*}[]
\centering
\begin{tabular}{|c|c|c|cccc|cccc|}
\hline
\multirow{2}{*}{\textbf{World}} &
  \multirow{2}{*}{\textbf{\begin{tabular}[c]{@{}c@{}}Number\\ of Runs\end{tabular}}} &
  \multirow{2}{*}{\textbf{\begin{tabular}[c]{@{}c@{}}Number \\ of Agents\end{tabular}}} &
  \multicolumn{4}{c|}{\textbf{CrowdSurfer}} &
  \multicolumn{4}{c|}{\textbf{Crowd-FM}} \\ \cline{4-11} 
 &
   &
   &
  \multicolumn{1}{c|}{\textbf{\begin{tabular}[c]{@{}c@{}}Trajectory\\ Length (m) $\downarrow$\end{tabular}}} &
  \multicolumn{1}{c|}{\textbf{\begin{tabular}[c]{@{}c@{}}Trajectory\\ Time (s) $\downarrow$\end{tabular}}} &
  \multicolumn{1}{c|}{\textbf{\begin{tabular}[c]{@{}c@{}}Velocity\\(m/s) $\uparrow$\end{tabular}}} &
  \textbf{\begin{tabular}[c]{@{}c@{}}Success\\ Rate $\uparrow$\end{tabular}} &
  \multicolumn{1}{c|}{\textbf{\begin{tabular}[c]{@{}c@{}}Trajectory\\ Length (m) $\downarrow$\end{tabular}}} &
  \multicolumn{1}{c|}{\textbf{\begin{tabular}[c]{@{}c@{}}Trajectory\\ Time (s) $\downarrow$\end{tabular}}} &
  \multicolumn{1}{c|}{\textbf{\begin{tabular}[c]{@{}c@{}}Velocity\\ (m/s) $\uparrow$\end{tabular}}} &
  \textbf{\begin{tabular}[c]{@{}c@{}}Success\\ Rate $\uparrow$\end{tabular}} \\ \hline
Cumberland &
  15 &
  \multirow{3}{*}{35} &
  \multicolumn{1}{c|}{18.82} &
  \multicolumn{1}{c|}{57.13} &
  \multicolumn{1}{c|}{0.52} &
  0.73 &
  \multicolumn{1}{c|}{\cellcolor{green!25}\textbf{16.18}} &
  \multicolumn{1}{c|}{\cellcolor{green!25}\textbf{36.49}} &
  \multicolumn{1}{c|}{\cellcolor{green!25}\textbf{0.71}} &
  \cellcolor{green!25}\textbf{0.87} \\ \cline{1-2} \cline{4-11} 
Lobby &
  18 &
   &
  \multicolumn{1}{c|}{13.87} &
  \multicolumn{1}{c|}{32.85} &
  \multicolumn{1}{c|}{0.61} &
  \cellcolor{green!25}\textbf{0.94} &
  \multicolumn{1}{c|}{\cellcolor{green!25}\textbf{13.76}} &
  \multicolumn{1}{c|}{\cellcolor{green!25}\textbf{25.42}} &
  \multicolumn{1}{c|}{\cellcolor{green!25}\textbf{0.72}} &
  0.89 \\ \cline{1-2} \cline{4-11} 
Freiburg &
  15 &
   &
  \multicolumn{1}{c|}{15.18} &
  \multicolumn{1}{c|}{44.71} &
  \multicolumn{1}{c|}{0.53} &
  0.73 &
  \multicolumn{1}{c|}{\cellcolor{green!25}\textbf{14.12}} &
  \multicolumn{1}{c|}{\cellcolor{green!25}\textbf{28.7}} &
  \multicolumn{1}{c|}{\cellcolor{green!25}\textbf{0.66}} &
  \cellcolor{green!25}\textbf{0.8} \\ \hline
Cumberland &
  15 &
  \multirow{2}{*}{45} &
  \multicolumn{1}{c|}{17.46} &
  \multicolumn{1}{c|}{85.81} &
  \multicolumn{1}{c|}{0.44} &
  0.67 &
  \multicolumn{1}{c|}{\cellcolor{green!25}\textbf{17.38}} &
  \multicolumn{1}{c|}{\cellcolor{green!25}\textbf{55.13}} &
  \multicolumn{1}{c|}{\cellcolor{green!25}\textbf{0.55}} &
  \cellcolor{green!25}\textbf{0.8} \\ \cline{1-2} \cline{4-11} 
Freiburg &
  15 &
   &
  \multicolumn{1}{c|}{14.24} &
  \multicolumn{1}{c|}{63.43} &
  \multicolumn{1}{c|}{\cellcolor{green!25}\textbf{0.56}} &
  0.6 &
  \multicolumn{1}{c|}{\cellcolor{green!25}\textbf{12.17}} &
  \multicolumn{1}{c|}{\cellcolor{green!25}\textbf{59.41}} &
  \multicolumn{1}{c|}{0.46} &
  \cellcolor{green!25}\textbf{0.73} \\ \hline
\end{tabular}
\vspace{-1mm}
\caption{\footnotesize{Performance comparison between Crowd-FM and Crowdsurfer. The metrics in the table are an average over the successful runs.}}
\label{crowdfm-v-crowdsurfer}
\vspace{-2mm}
\end{table*}

\begin{table*}[htbp]
\centering
\renewcommand{\arraystretch}{1.2} 
\setlength{\tabcolsep}{4pt}  
\begin{tabular}{|c|c|cccc|cccc|}
\hline
\multirow{2}{*}{\textbf{World}} &
  \multirow{2}{*}{\textbf{\begin{tabular}[c]{@{}c@{}}Number\\ of Runs\end{tabular}}} &
  \multicolumn{4}{c|}{\textbf{DRL-VO}} &
  \multicolumn{4}{c|}{\textbf{Vanilla CFM}} \\ \cline{3-10} 
 &
   &
\multicolumn{1}{c|}{\textbf{\begin{tabular}[c]{@{}c@{}}Trajectory\\ Length (m)\end{tabular}} $\downarrow$} &
\multicolumn{1}{c|}{\textbf{\begin{tabular}[c]{@{}c@{}}Trajectory\\ Time (s)\end{tabular}} $\downarrow$} &
\multicolumn{1}{c|}{\textbf{\begin{tabular}[c]{@{}c@{}}Velocity\\ (m/s)\end{tabular}} $\uparrow$} &
\multicolumn{1}{c|}{\textbf{\begin{tabular}[c]{@{}c@{}}Success\\ Rate\end{tabular}} $\uparrow$} &
\multicolumn{1}{c|}{\textbf{\begin{tabular}[c]{@{}c@{}}Trajectory\\ Length (m)\end{tabular}} $\uparrow$} &
\multicolumn{1}{c|}{\textbf{\begin{tabular}[c]{@{}c@{}}Trajectory\\ Time (s)\end{tabular}} $\downarrow$} &
\multicolumn{1}{c|}{\textbf{\begin{tabular}[c]{@{}c@{}}Velocity\\ (m/s)\end{tabular}} $\uparrow$} &
\multicolumn{1}{c|}{\textbf{\begin{tabular}[c]{@{}c@{}}Success\\ Rate\end{tabular}} $\uparrow$} \\ \hline

Cumberland &
  15 &
  \multicolumn{1}{c|}{18.47} &
  \multicolumn{1}{c|}{55.95} &
  \multicolumn{1}{c|}{0.40} &
  0.40 &
  \multicolumn{1}{c|}{\cellcolor{green!25}\textbf{16.58}} &
  \multicolumn{1}{c|}{\cellcolor{green!25}\textbf{54.42}} &
  \multicolumn{1}{c|}{\cellcolor{green!25}\textbf{0.48}} &
  \cellcolor{green!25}\textbf{0.67} \\ \hline
Lobby &
  18 &
  \multicolumn{1}{c|}{\cellcolor{green!25}\textbf{13.15}} &
  \multicolumn{1}{c|}{39.89} &
  \multicolumn{1}{c|}{0.39} &
  \textbf{0.61} &
  \multicolumn{1}{c|}{13.61} &
  \multicolumn{1}{c|}{\cellcolor{green!25}\textbf{36.99}} &
  \multicolumn{1}{c|}{\cellcolor{green!25}\textbf{0.48}} &
  \textbf{0.61} \\ \hline
Freiburg &
  15 &
  \multicolumn{1}{c|}{14.18} &
  \multicolumn{1}{c|}{44.50} &
  \multicolumn{1}{c|}{0.41} &
  0.53 &
  \multicolumn{1}{c|}{\cellcolor{green!25}\textbf{13.43}} &
  \multicolumn{1}{c|}{\cellcolor{green!25}\textbf{34.45}} &
  \multicolumn{1}{c|}{\cellcolor{green!25}\textbf{0.51}} &
  \cellcolor{green!25}\textbf{0.6} \\ \hline
\end{tabular}
\vspace{-1mm}
\caption{\footnotesize{Performance comparison between the Vanilla CFM(with Cost Guidance) and DRL-VO planner with 35 agents. Metrics for Trajectory Length, Time, and Velocity are averaged over all successful runs. The results demonstrate that CFM models collision avoidance more effectively than DRL-VO, achieving a higher success rate, shorter paths, higher average velocity, and lower trajectory times.}}
\label{tab:vanilla_results}
\vspace{-2mm}
\end{table*}




\begin{table}[]
\centering
\begin{tabular}{|c|c|}
\hline
\textbf{Vanilla CFM} & \textbf{Success Rates $\uparrow$} \\ \hline
w/o Cost Guidance    & 0.57                   \\ \hline
w/ Cost Guidance     & \cellcolor{green!25}\textbf{0.67}                 \\ \hline
\end{tabular}
\vspace{-1mm}
\caption{\footnotesize{We test the impact of Implicit Cost-Guidance on the performance of Vanilla-CFM over 10 runs with K=5 integration steps. The increase in the success rate indicates the usefulness of the guidance.}}
\label{tab:guidance}
\end{table}
 
\subsubsection{\textbf{Effect of Inference-time Cost Guidance}}

\begin{figure}[ht]
    \vspace{-0.2cm}
    \centering
    \includegraphics[width=0.9\linewidth]{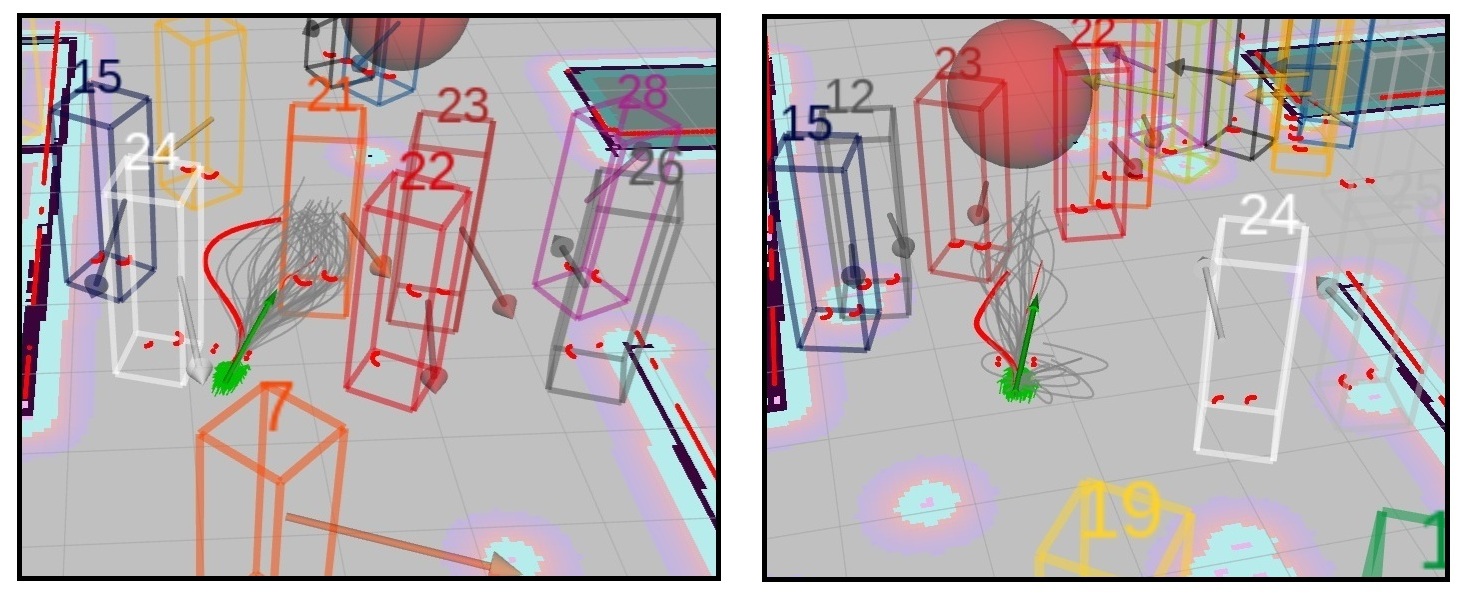}
    \vspace{-0.2cm}
    \caption{\footnotesize{\textit{Effect of Inference-time Cost Guidance}: Left image shows the trajectories generated by Crowd-FM \textit{with} Cost Guidance. The right image shows the trajectories generated by Crowd-FM \textit{without} Cost Guidance. Trajectories generated using collision cost guidance are found to be more controlled near obstacles than the trajectories without it. (Tests done without Optimizer Refinement)}}
    \label{fig_px}
    \vspace{-0.45cm}
\end{figure}

One of the two ways we employ inference-time refinement of CFM-generated trajectories is the Inference-time Cost Guidance.
We test this property of Crowd-FM by disabling Optimizer refinement on the Cumberland and Freiburg environments in Pedsim with 25 agents for 10 trials.
Table \ref{tab:guidance} shows a clear improvement in Success rates for the collision-cost guided Flow objective compared to the un-guided one. Additionally, we present qualitative results for the same in Fig. \ref{fig_px}.

\subsubsection{\textbf{Effect of the Learned Scoring Function}}

To validate that the proposed Scoring Function improves the human-likeness of robot trajectories, 
we introduce a quantitative metric termed the \textit{Human-Likeness Points(HLP)}. 
For a candidate trajectory $P_c$ and the corresponding expert (human) trajectory $P_h$, we define HLP as the average point-wise deviation over a time horizon $T$:


\begin{table}[]
\vspace{-3mm}
\centering
\begin{tabular}{|c|ccc|}
\hline
\textbf{Method} & \multicolumn{3}{c|}{\textbf{Success Rate $\uparrow$}} \\
\cline{2-4}
& \textbf{Cumberland} & \textbf{Lobby} & \textbf{Freiburg} \\
\hline
Cost Function & 0.80 & 0.83 & \textbf{0.80} \\
\hline
Score Function (Ours) & \cellcolor{green!25}\textbf{0.87} & \cellcolor{green!25}\textbf{0.89} & \textbf{0.80} \\
\hline
\end{tabular}
\vspace{-1mm}
\caption{\footnotesize{Comparison of predefined cost-based planning and our learned Scoring Function across different environments.}}
\label{tab:success_results}
\vspace{-0.7cm}
\end{table}

\[
    \text{HLP}(P_c, P_h) = \frac{1}{T} \sum_{t=1}^{T} \lVert P_c(t) - P_h(t) \rVert_2.
\]

\begin{figure}[ht]
    \centering
    \includegraphics[width=0.9\linewidth]{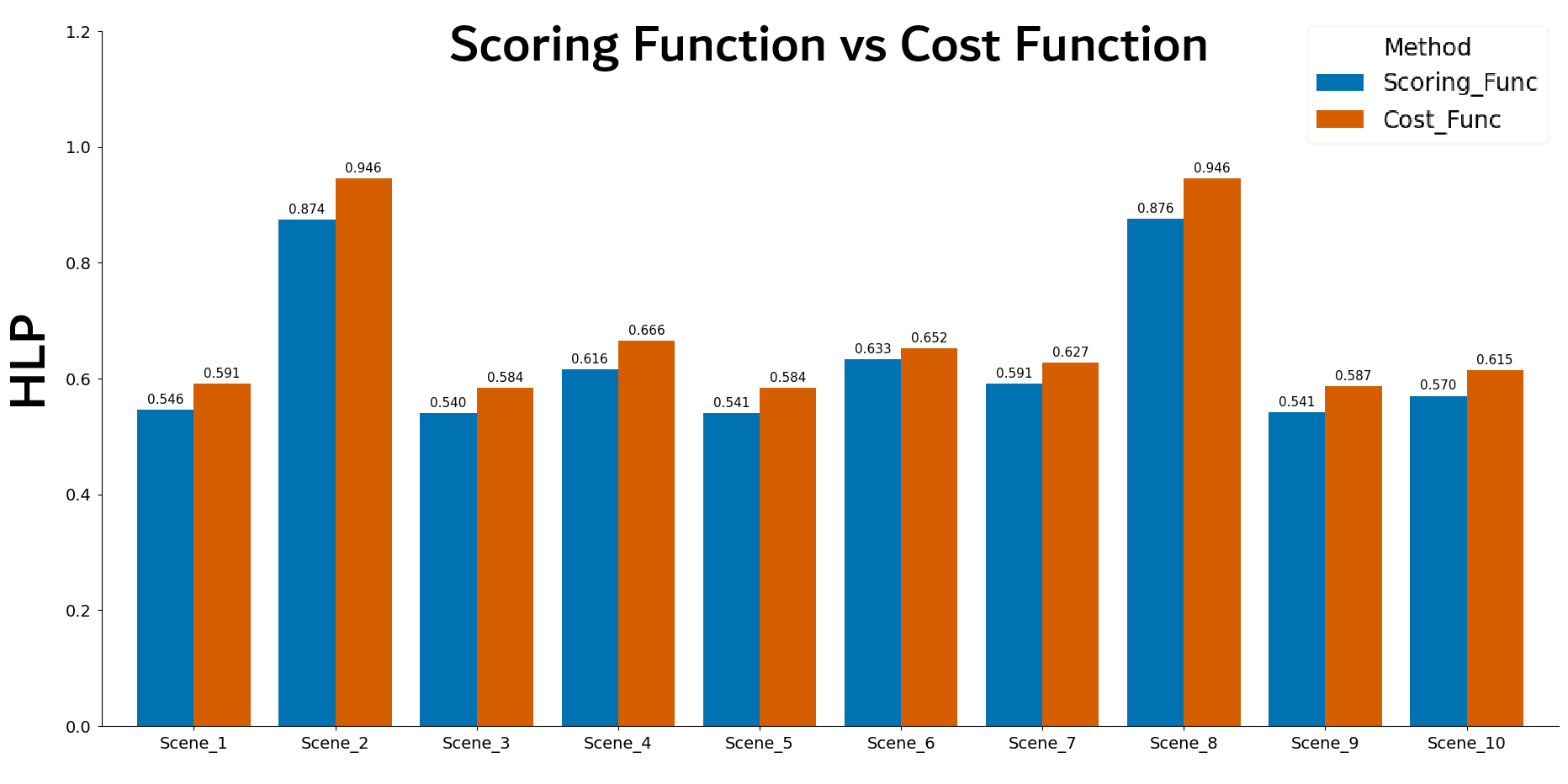}
    \caption{\footnotesize{Grouped Bar Plot showing the HLPs for Scoring Function selected Trajectories(\textcolor{blue}{BLUE}) and Cost-Function selected Trajectories(\textcolor{orange}{ORANGE}).}}
    \label{fig_bar}
    \vspace{-0.5cm}
\end{figure}

Lower HLP values indicate that the chosen trajectory is closer to the expert demonstration and thus more human-like. We compare trajectories selected by our learned Scoring Function against those chosen by predefined, hand-tuned cost functions that primarily optimize for collision avoidance and smoothness. These tests were done in an open-loop setting to take into account the expert human demonstrations. A custom wheelchair prototype was teleoperated in indoor environments amid dense crowds to create the test dataset. We evaluated HLPs for both methods on 10 such scenes and provided our results as a Grouped Bar Chart in Fig. \ref{fig_bar}.

\vspace{1.5mm}

Evidently, our Scoring Function consistently reduces the HLP across all environments, 
indicating that the selected trajectories are closer to expert data. 
At the same time, success rates in Table \ref{tab:success_results} improve over different closed-loop environments. 
This demonstrates that the scoring mechanism not only preserves safety but also aligns robot motion more closely with human navigation patterns.

\subsubsection{\textbf{Runtime Analysis}} \label{sec:runtime}
To evaluate real-time feasibility, we benchmarked the Crowd-FM pipeline on a laptop with an NVIDIA RTX 3060 GPU. The entire planning loop, including data processing and trajectory generation, takes $\sim$75ms. A detailed breakdown shows that the CFM model generates the initial trajectory batch in $\sim$45ms, and the Learned Scoring Function selects the optimal candidate in $\sim$5ms. The remaining time is utilized by the Optimizer Refinement step ($\sim$25ms), which ensures dynamic feasibility. Performances are naturally expected to improve with better GPUs. Crowd-FM's high computational efficiency maintains a consistent control rate suitable for dense crowd navigation.

\subsection{Real-World Experiments}

\begin{figure}[h]
    \vspace{-0.2cm}
    \centering
    \includegraphics[width=0.9\linewidth]{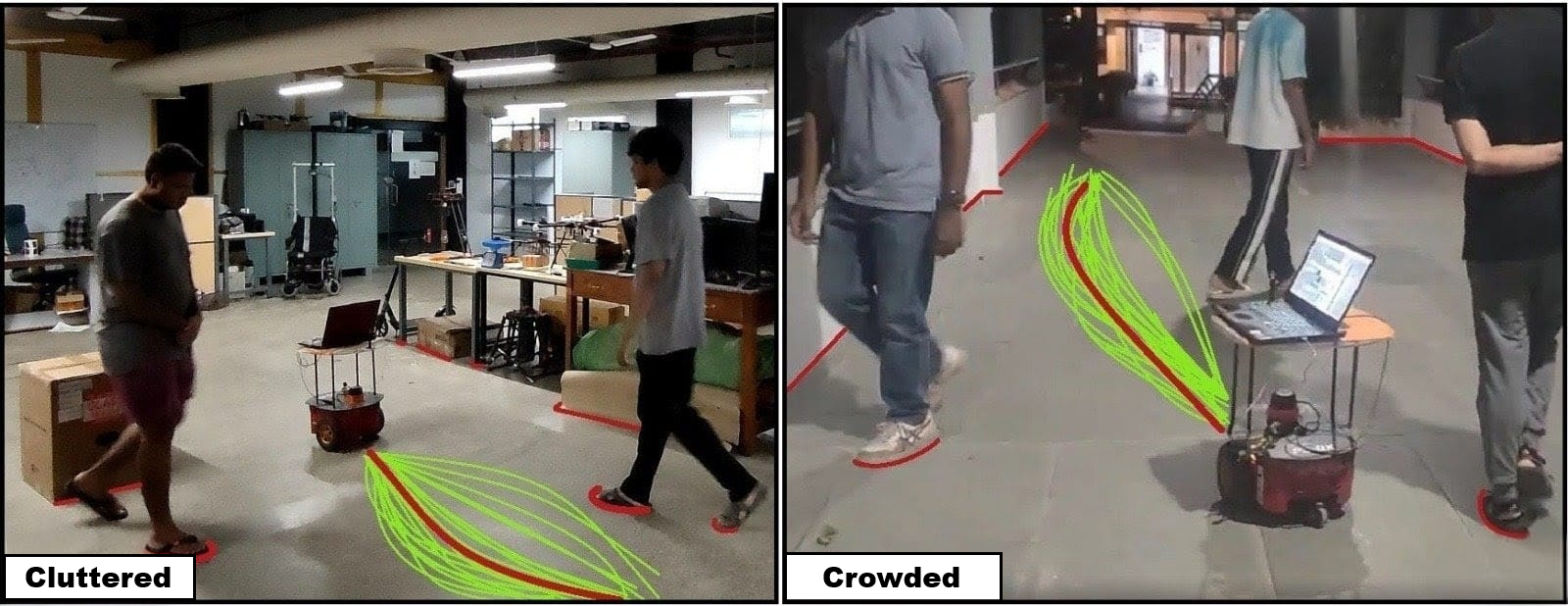}
    \caption{\footnotesize{Crowd-FM in action on a \textbf{P3-DX robot} in two distinct scenarios - a cluttered environment with high static obstacle density(lab), and a crowded environment with high dynamic obstacle density(academic block).
    }}
    \label{real-world}
    \vspace{-0.5cm}
\end{figure}

We validated Crowd-FM in real-world scenarios on a Pioneer 3-DX mobile robot. The robot was equipped with an RPLiDAR 2D scanner and utilized a LegTracker\cite{legtracker} module to estimate dynamic obstacle velocities and positions. The experiments were conducted in our lab and in the academic block, where our method consistently demonstrated reasonable success rates in densely populated environments. In these complex settings, the system achieved an overall success rate of 17 runs out of 20 total trials. Fig.~\ref{real-world} shows stills from two such representative testing scenarios.

\vspace{0.1cm}
\section{CONCLUSION AND FUTURE WORK}
\vspace{-0.1cm}
This work introduces Crowd-FM, a trajectory-planning framework that utilizes Conditional Flow Matching to learn expressive distributions over Bernstein polynomial control points. Unlike prior generative approaches, which have discrete latent bottlenecks(VQ-VAEs) or are computationally expensive(Diffusion), Crowd-FM demonstrates that flow-based transport can capture diverse and smooth navigation, while remaining efficient at inference. A lightweight projection optimization step further refines the candidate trajectories, boosting success by $\sim$20\%, without resorting to a slower complete trajectory optimization. We validate the approach both in simulation and real-world settings, highlighting that smooth, non-freezing, and human-like trajectories can be rapidly generated in dense crowds.

An essential future direction is to expand the dataset with more heterogeneous scenarios, allowing us to capture richer behaviors and test the scoring function under different crowd dynamics. We also see CFM's potential in learning social cues and scene semantics through its integration with complementary planning modules, such as human motion prediction models, moving towards social navigation.

\addtolength{\textheight}{-12cm}   









\section{Acknowledgment}
The authors would like to acknowledge the use of large language models (Gemini, ChatGPT) and writing enhancement tools (Grammarly) solely for refining the grammatical structure and the linguistic flow of this manuscript.

\bibliographystyle{ieeetr}
\bibliography{bibliography}

\end{document}